\documentclass[11pt]{amsproc}

\usepackage{times}
\usepackage{latexsym}
\usepackage{hyperref}
\usepackage{booktabs}
\usepackage{import}
\usepackage{tikz}
\usepackage{geometry}
\usepackage[final]{pdfpages}
\usepackage{blindtext}
\usepackage{setspace}
\usepackage{color, colortbl}

\definecolor{G1}{gray}{1.00}
\definecolor{G2}{gray}{0.96}
\definecolor{G3}{gray}{0.92}
\definecolor{G4}{gray}{0.88}

\usepackage[T1]{fontenc}
\usepackage[utf8]{inputenc}

\usepackage{microtype}

\title{Argumentation Element Annotation Modeling using XLNet}
\author{Christopher Ormerod, Amy Burkhardt, Mackenzie Young, and Sue Lottridge}

\date{November 10th 2023}

\begin{document}

\onehalfspacing

\maketitle

\begin{abstract}
This study demonstrates the effectiveness of XLNet, a transformer-based language model, for annotating argumentative elements in persuasive essays. XLNet's architecture incorporates a recurrent mechanism that allows it to model long-term dependencies in lengthy texts. Fine-tuned XLNet models were applied to three datasets annotated with different schemes - a proprietary dataset using the Annotations for Revisions and Reflections on Writing (ARROW) scheme, the PERSUADE corpus, and the Argument Annotated Essays (AAE) dataset. The XLNet models achieved strong performance across all datasets, even surpassing human agreement levels in some cases. This shows XLNet capably handles diverse annotation schemes and lengthy essays. Comparisons between the model outputs on different datasets also revealed insights into the relationships between the annotation tags. Overall, XLNet's strong performance on modeling argumentative structures across diverse datasets highlights its suitability for providing automated feedback on essay organization.
\end{abstract}

\section{Introduction}

Persuasive essays aim to influence the reader's viewpoint on an issue through compelling arguments. Crafting persuasive arguments is a crucial skill for decision-making and problem-solving. A persuasive essay is structured to logically present evidence and reasoning that supports the author's position. By analyzing the components of an essay, including the main claims, supporting claims, and corroborating evidence, we can understand its argumentative structure \cite{stab_annotating_2014}. Recent research shows that highlighting these argument elements for students in their essays helps improve their persuasive writing abilities \cite{butler_investigating_2011}.  Integrating models that can automatically annotate argument components into Automated Writing Evaluation systems would enable providing students valuable feedback on the organizational structure of their persuasive essays.

Discourse analysis involves studying how language constructs arguments and conveys persuasive messages. To facilitate discourse analysis research, many datasets and annotation schemes have been developed. Some schemes focus on the hierarchical relationships between discourse units like paragraphs and sentences, as well as the rhetorical relations connecting them. Notable datasets annotated this way include the Rhetorical Structure Theory (RST) Treebank \cite{prasad_penn_nodate}, the Argument Annotated Essays (AAE) dataset \cite{stab_annotating_2014}, and the arg-microtext corpus \cite{peldszus_joint_2015}. Other schemes partition the introduction, conclusion, and body paragraphs of essays into functional organizational components. One such scheme was used to annotate the PERSUADE corpus \cite{crossley_persuasive_2022}. The key difference is that some annotation schemes capture the hierarchical and rhetorical connections between discourse units, while others delineate the functional roles of organizational components within essays. Both approaches provide insights into the structure and argumentation of persuasive writing.

The primary goal of this study is to demonstrate the exceptional suitability of pretrained XLNet models for accurately modeling annotation schemes that identify argument components in essays. To showcase XLNet's performance, we fine-tuned two versions of XLNet on three datasets annotated with different schemes. The first is a proprietary dataset annotated using the "Annotations for Revisions and Reflections On Writing" (ARROW) scheme \cite{ARROW}designed for this study. The second dataset is the PERSUADE corpus annotated with a similar scheme \cite{crossley_persuasive_2022}. The third is the smaller argument-annotated essay (AAE) dataset which highlights structural argument properties \cite{stab_annotating_2014}. The models derived from these datasets allow approximating relationships between the annotation tags across schemes and enable more detailed analysis of essay argument structure. Fine-tuning XLNet on these diverse corpora annotated with different schemes highlights its versatility for accurately modeling argument components, supporting its integration into automated writing evaluation systems.

XLNet is well-suited for modeling argument annotation schemes due to its ability to capture long-term dependencies without length restrictions. Most transformer-based language models like BERT, GPT, and their variants have an inherent input length limit of 512 tokens, based on the original transformer design by Vaswani et al. \cite{vaswani_attention_2017}. While adequate for many NLP benchmarks \cite{wang_glue_2019, wang_superglue_2020}, this poses challenges for modeling essays which often exceed 512 tokens. Accurately annotating argument components relies on capturing long-range dependencies across an entire essay. However, XLNet uses a novel recurrent transformer formulation allowing it to model dependencies without a fixed length limit \cite{yang_xlnet_2019}. This makes XLNet naturally adept at handling full-length essays and modeling annotation schemes that rely on global document context. Its recurrent attention mechanism handles long essays as effortlessly as short texts. This key advantage underpins XLNet's effectiveness at argument annotation modeling.

The key feature enabling XLNet to model long essays is its recurrent transformer architecture \cite{dai_transformer-xl_2019}.
Although XLNet is trained on 512-token segments, the recurrence mechanism provides a substantially longer relative effective context length \footnote{See Appendix A of \cite{dai_transformer-xl_2019} to see the precise definition of relative effective context length.}. This allows it to capture dependencies well beyond a single segment. Two pretrained XLNet versions are available \cite{wolf_huggingfaces_2020}: a 110 million parameter base model and a 330 million parameter large model. We aim to showcase the strong performance of both models on our three datasets. The only minor limitation when handling very long inputs is the memory required during training. To mitigate this, we specify memory-saving optimizations in our training procedures. Overall, XLNet's recurrent transformer formulation, with its long effective context length, makes it uniquely capable of handling full-length essays and learning argument annotation schemes.

The article is structured in the following manner: Section 2 outlines the annotation schemes used in this study, the data used, and modeling specifics. In Section 3, we evaluate our models performance and consider relations between the annotation schemes. Finally, in Section 4, we discuss our findings and suggest areas for future research.

\section{Methods}

In this section, we briefly discuss the three annotation schemes used in this study; the ARROW scheme, the scheme used to annotate the PERSUADE corpus, and the scheme used to annotate the AAE dataset. In the modeling sections, we specify how the XLNet models were applied, which includes how the inputs and targets are defined. 

\subsection{Annotation Schemes}

Each of the annotation schemes are designed for slightly different purposes. We will discuss these differences and how these differences may be relevant to standards being assessed.

\subsubsection{The ARROW Annotation Scheme}

The ARROW annotation scheme was constructed in consultation with experts  to provide feedback that aligns with standards in the assessment of source dependent persuasive essays. This scheme defines seven annotation tags corresponding to argumentative elements. Annotations were applied at a sentence level where sentences were defined by sentence-ending punctuation or paragraph boundaries. 

The seven annotation tags applied by human raters were as follows: 
\begin{itemize}
\item{{\bf Introduction (I1):} A plan of the argument, such as listing subtopics, the use of rhetorical devices to establish context, and attention-grabbing devices. This should not include a well-defined controlling idea sentence.}
\item{{\bf Controlling Idea (I2):} Any sentence in which the author has a claim stating the author's stance on an issue.}
\item{{\bf Evidence (E1):} Sentences that include citations, quotations, and or data from sources, or a paraphrased version of another source.}
\item{{\bf Elaboration (E2):} Sentences containing general arguments, reasons, and commentary that support claims, any sub-claims, and rhetorical devices to enhance arguments. Also any rebuttal for an opposing position. }
\item{{\bf Opposing Position (O):} Sentences that include any acknowledgment of an opposing position not in the introduction or conclusion, or any sentences that explore the stated opposing position.}
\item{{\bf Conclusion (C):} Sentences that summarize the evidence and elaboration. It should not include any new ideas.}
\item{{\bf Transitions (T):} Sentences with no new information intending to create coherence or structure. These include sentences at the beginning of a paragraph that signal what is coming next or at the end of a paragraph reiterating a claim.}
\end{itemize}
This annotation scheme was designed to align with many state standards for argumentative essay writing for grades 6 to 8. We define an alignment between a standard and an annotation tag to mean that the presence of a particular annotation tag can be used as evidence that the student has met a standard.

While it is important to know that the set of standards for argumentative essay writing varies from state to state, many commonalities arise. Most standards specify, in one form or another, that students are to clearly introduce claims and opposing claims. In the above sense, the standards of this form align with (I2) and (O). Secondly, students are typically required to support claims with logical reasoning with relevant evidence using accurate, credible sources and demonstrating an understanding of the topic or text, which aligns with (E1) and (E2). Standards that require the provision of a concluding statement that follows from and supports the argument presented align with (C). Many standards also have some reference to organizing an essay in a coherent manner, but also the use of phrases to clarify the relationships between claims and reasons. Such standards align with (I1), (C), and (T). 

In training the hand-scoring team, when two or more of these tags apply to the same sentence, the annotator was instructed to apply an automatic resolution process. This process is specified by a hierarchy in which the annotator should apply the first annotation that applies in the following ordering:  
\[
I2 \rightarrow O \rightarrow E1 \rightarrow E2 \rightarrow T
\]
Note that (I1) and (C) are not included in this hierarchy. These serve different purposes than the essay's body and are considered separate entities of an organization that help frame and structure the text as a whole. 

In this project, we curated a collection of 18,000 essays from 9 different prompts ranging from grades 6 to 8 from one state on their summative online assessment program. Any empty or inappropriate responses were removed. From this collection, a random collection of approximately 15\% of all essays were annotated by two raters, firstly in order to gauge the inter-rater reliability, and secondly, as a means to control the quality of hand-scoring. We call the set of essays that receive two scores the validation sample. Table \ref{tab:data_summary} presents the grade for each prompt, the total number of responses that were annotated, the size of the validation sample, and the average length given by the number of words in each essay.

\begin{table}[!ht]
    \centering
    \begin{tabular}{c|c c c c c c c } \toprule
\rowcolor{G4}        Prompt & Grade & Total & Val & Avg Len\\ \midrule 
\rowcolor{G2}        \# 1 & 7 & 1,955 & 298 & 448\\
\rowcolor{G1}        \# 2 & 7 & 1,940 & 283 & 446\\
\rowcolor{G2}        \# 3 & 8 & 1,937 & 304 & 484\\
\rowcolor{G1}        \# 4 & 8 & 1,955 & 302 & 512\\
\rowcolor{G2}        \# 5& 8 & 1,950 & 294 & 480\\ \hline
\rowcolor{G1}        \# 6 & 7 & 1,953 & 297 & 449\\
\rowcolor{G2}        \# 7 & 6 & 1,929 & 261 & 397\\
\rowcolor{G1}        \# 8 & 6 & 1,932 & 298 & 384\\
\rowcolor{G2}        \# 9 & 6 & 1,933 & 283 & 401\\ \midrule
\rowcolor{G3}        Total &  & 17,484 & 2,619 &  \\ \bottomrule
    \end{tabular}
    \caption{A list of prompts with the number of training samples, validation samples, and average number of words in each essay for each prompt.}
    \label{tab:data_summary}
\end{table}

The validation sample has a well-defined resolution tag for each sentence; if two raters agree, then it is the agreed-upon tag, and if the raters disagree, we treat this as a case in which two or more tags apply, hence, the automatically resolved tag is defined by the hierarchy above. This means we are able to determine whether the agreement between the resolved score and our model is greater than the agreement between two human raters. The metric we use to define agreement is the Cohen's kappa statistic \cite{cohen_psychological_1996}, given by
\begin{equation}
\kappa = \dfrac{p_o - p_e}{1-p_e},
\end{equation}
where $p_o$ is the observed agreement and $p_e$ is the expected agreement. This can be done at an individual tag level where our corpus of essays are treated as collection of sentences. This mimics typical criteria used in automated scoring \cite{williamson_framework_2012}. Each sentence in each essay is considered an independent tag, hence, when considering agreements in the validation process, we consider the sequence of all sentences appearing in the validation sample. The IRR metrics for each prompt and each tag is presented in Table \ref{tab:irr_handscoring}.

\begin{table}[!ht]
\small
    \centering
    \begin{tabular}{c|c c c c c c c} \toprule
\rowcolor{G4}         & I1 & I2 & E1 & E2 & O & C & T \\ \midrule 
\rowcolor{G2}        1  & 0.81 & 0.79 & 0.65 & 0.63 & 0.56 & 0.82 & 0.42\\
\rowcolor{G1}        2  & 0.71 & 0.77 & 0.65 & 0.61 & 0.46 & 0.80 & 0.37\\
\rowcolor{G2}        3  & 0.78 & 0.72 & 0.56 & 0.59 & 0.42 & 0.82 & 0.33\\
\rowcolor{G1}        4  & 0.75 & 0.78 & 0.66 & 0.64 & 0.60 & 0.78 & 0.41\\
\rowcolor{G2}        5  & 0.75 & 0.80 & 0.68 & 0.66 & 0.46 & 0.86 & 0.49\\ \hline
\rowcolor{G1}        6  & 0.75 & 0.72 & 0.63 & 0.64 & 0.58 & 0.85 & 0.50\\
\rowcolor{G2}        7  & 0.69 & 0.66 & 0.63 & 0.57 & 0.26 & 0.75 & 0.25\\
\rowcolor{G1}        8  & 0.77 & 0.71 & 0.48 & 0.51 & 0.26 & 0.84 & 0.50\\
\rowcolor{G2}        9  & 0.68 & 0.58 & 0.58 & 0.57 & 0.09 & 0.77 & 0.32\\ \midrule 
\rowcolor{G3}        Avg & 0.76 & 0.74 & 0.63 & 0.62 & 0.48 & 0.83 & 0.43\\\bottomrule
    \end{tabular}
    \caption{The kappa statistics indicate the inter-annotator reliability levels between two raters on the annotated corpus.}.
    \label{tab:irr_handscoring}
\end{table}

Highly imbalanced classes posed an additional challenge to modeling. The most frequently used tag by far was the Elaboration (E2) tag which constituted almost half the sentences in the dataset. This was followed by Introduction (I1), Conclusion (C), and Evidence (E1), which were all sufficiently well-represented for modeling purposes, however, the remaining tags do not appear with a high frequency. We also note that Opposing Position (O) is an argumentative technique that is emphasized more in grades 7 and 8, hence, they are very infrequently used in grade 6 essays. This is reflected in the very low kappa values for Opposing Position for prompts 7,8, and 9. 

\begin{table}
\centering
\begin{tabular}{l r r } \toprule
\rowcolor{G4}Class & \# sentences & percentage \\ \midrule
\rowcolor{G2}Introduction & 48,526 & 12.8\% \\
\rowcolor{G1}Controlling Idea & 19,520 & 5.1\%  \\
\rowcolor{G2}Evidence & 59,338 & 15.6\%  \\
\rowcolor{G1}Elaboration & 182,920 & 48.1\%  \\
\rowcolor{G2}Opposing Position & 20,091 & 5.3\% \\
\rowcolor{G1}Conclusion & 43,177 & 11.4\% \\ 
\rowcolor{G2}Transitions & 5,114 & 1.3\% \\ \midrule
\rowcolor{G3}Total & 380,214 & 99.6\% \\ \bottomrule
\end{tabular}
\caption{The percentage of sentences in the dataset designated to each argumentation element. The remaining 0.4\% of sentences were given no annotation.}
\end{table}

In addition to the corpus of hand-scored data, one of the goals of this study was to utilize a large corpus of over 250k essays from the same administrative platform were provided to improve modeling in a semi-supervised manner. This corpus included essays from 47 prompts from the same administration where none of prompts were included in the original data. 

\subsubsection{The PERSUADE Corpus}

The PERSUADE corpus was only recently introduced on the Kaggle website as part of a Feedback Prize\footnote{https://www.kaggle.com/c/feedback-prize-2021}. This dataset is a large open-source corpus of essays with annotations that outline argumentative components and relations, a host of demographic data, and holistic essay scores. The extended demographic information and holistic scores are provided separately\footnote{The corpus is available for download at https://github.com/scrosseye/persuade\_corpus\_2.0}. The utility of this dataset modeling for use in AWE systems, and bias, and other quantitative analysis is remarkable. In terms of length, these essays range from 150 to 2000 words with an average of approximately 400 words per essay, making this an ideal corpus to test the long term dependencies of the XLNet model.

The definition of the annotation tags applied in the PERSUADE corpus is given by:
\begin{itemize}
\item {\bf Lead (L):} An introduction that begins with a statistic, a quotation, a description, or some other device to grab the reader’s attention and point toward the thesis. 
\item  {\bf Position (P):} An opinion or conclusion on the main question. 
\item {\bf Claim (C1):} A claim that supports the position. 
\item {\bf Counterclaim (C2):} A claim that refutes another claim or gives an opposing reason to the position. 
\item {\bf Rebuttal (R):} A claim that refutes a counterclaim. 
\item {\bf Evidence (E):} Ideas or examples that support claims, counterclaims, rebuttals, or the position. 
\item {\bf Concluding Statement (C3):} A concluding statement that restates the position and claims.
\end{itemize}
The annotations in the PERSUADE corpus align well with the standards for persuasive essay writing in many states. The Lead (L) and Concluding Statement (C3) relate to the coherent organization of an essay, while the Position (P), Claims (C1), Counterclaims (C2), and Rebuttals (R) strongly align with the standard pertaining to the statement of all claims and opposing claims. The Evidence (E) tag includes all ideas or examples to support the claims, which is a distinction from the ARROW scheme, where most ideas would be considered Elaboration (E2). In the context of both source dependent and independent essays, the Evidence (E) tag aligns well with standards requiring either logical reasoning or accurate, credible sources.

The PERSUADE corpus contains a total of 25,996 essays collected from students between grades 6 and 11. The training set consists of 15,594 essays while the test set contains 10,402 essays. The essay prompts, the grades assigned for these prompts, and the proportion of essays from each particular prompt are shown in Table \ref{tab:persuade-prompts}. The distribution of grades between the dataset for the ARROW scheme and the PERSUADE corpus is important because the standards applied to each grade change in a significant way between grades 6 and 7. For argumentative essay writing, the use of opposing positions and counterclaims only appears prominently in the standards for grade 7 and above. Unlike the ARROW scheme, Rebuttal (R) is distinguished from Elaboration (E2), which associates counterarguments with two annotation tags instead of one, namely Counterclaim (C2) and Rebuttal (R). This separation may also imply that the PERSUADE scheme is actually more appropriate for the annotation of essays for higher grades than the ARROW scheme.

\begin{table}
    \centering
    \begin{tabular}{l|l l r r } \toprule
\rowcolor{G4}Prompt & Grade & Text & Train & Test \\ \midrule
\rowcolor{G1}"A Cowboy Who Rode the Waves"         & 6   & Dep. & 4.4\% & 6.6\%\\
\rowcolor{G2}Cell phones at school                 & 8   & Ind. & 5.3\% & 8.0\%\\
\rowcolor{G1}Community service                     & 8   & Ind. & 4.9\% & 7.5\%\\
\rowcolor{G2}Grades for extracurricular activities & 8   & Ind. & 5.2\% & 7.8\%\\
\rowcolor{G1}Mandatory extracurricular activities  & 8   & Ind. & 5.4\% & 8.0\%\\
\rowcolor{G2}Seeking multiple opinions             & 8   & Ind. & 9.9\% & 0\%\\
\rowcolor{G1}The Face on Mars                      & 8   & Dep. & 5.2\% & 7.4\%\\
\rowcolor{G2}Does the electoral college work?      & 9   & Dep. & 11.6\% & 2.2\%\\
\rowcolor{G1}Car-free cities                       & 10  & Dep. & 6.3\% & 9.4\%\\
\rowcolor{G2}Driverless cars                       & 10  & Dep. & 8.9\% & 4.8\%\\
\rowcolor{G1}Exploring Venus                       & 10  & Dep. & 6.0\% & 8.9\%\\
\rowcolor{G2}Facial action coding system           & 10  & Dep. & 7.1\% & 10.2\%\\
\rowcolor{G1}Distance learning                     & 11  & Ind. & 9.6\% & 6.3\%\\
\rowcolor{G2}Summer projects                       & 11  & Ind. & 5.6\% & 8.4\%\\
\rowcolor{G1}Phones and driving                    & N/A & Ind. & 4.5\% & 4.5\%\\ \bottomrule
    \end{tabular}
    \caption{An enumeration of the various prompts, grades, whether they are dependent on a source text (Dep.) or are independent of a source text, and their representation in the train/test split. }
    \label{tab:persuade-prompts}
\end{table}

In terms of modeling, the distribution of annotation tags in the PERSUADE corpus, shown in Table \ref{tab:dist_persuade}, also poses some problems. The assignment of the Counterclaim (C2) and Rebuttal (R) tags are exceedingly rare. Furthermore, given the inter-rater reliability for opposing position (O) in the ARROW scheme was fairly low, we expect that agreement between two raters for the Counterclaim (C2) tag and Rebutal tag (R) to be fairly low as well. Unfortunately, one of the limitations of this corpus is that there is no way to determine the reliability of a tag. This also has an implication for the most appropriate metric to use. Instead of the Cohen's kappa statistic, we use the human assigned tags as the ground truth in calculating the F1 score, defined as
\begin{eqnarray}
F1 &=& \dfrac{2 P R}{P+R}\\
P &=& \dfrac{T_p}{T_p + F_p}\\
R &=& \dfrac{T_p}{T_p + F_n}
\end{eqnarray}
where $T_p$ is the number of true positives, $F_p$ is the number of false positives, and $F_n$ is the number of false negatives. Similar to the Cohen's kappa statistic, the F1 score takes the imbalanced classes intro consideration.

\begin{table}
    \centering
    \begin{tabular}{l |r  r | r r } \toprule
\rowcolor{G4}         &  \multicolumn{2}{c|}{Train} & \multicolumn{2}{c}{Test} \\
\rowcolor{G3}         & \# tokens & percentage &  \# tokens & percentage \\ \midrule 
\rowcolor{G1}        Lead                    & 483,365    &  7.4\%   & 292,982  &  7.2\% \\
\rowcolor{G2}        Position                & 281,334    &  4.3\%   & 183,642  &  4.5\% \\
\rowcolor{G1}        Claim                   & 874,631    & 13.4\%   & 569,585  & 14.0\% \\ 
\rowcolor{G2}        Counterclaim            & 139,816    &  2.1\%   & 87,724   &  2.1\% \\
\rowcolor{G1}        Rebuttal                & 121,839    &  1.9\%   & 83,428   &  2.0\% \\
\rowcolor{G2}        Evidence                & 3,535,759   & 54.2\%   & 2,194,660 & 53.8\% \\
\rowcolor{G1}        Concluding Statement    & 827,963    & 12.7\%   & 510,321  & 12.5\% \\
\rowcolor{G2}        None                    & 25,6108    &  3.9\%  & 158,650   &  3.9\% \\ \midrule
\rowcolor{G3}        Total & 6,520,815 & 100\% & 4,080,992 & 100\%\\ \bottomrule
    \end{tabular}
    \caption{A distribution of the various tags assigned to the tokens.}
    \label{tab:dist_persuade}
\end{table}

Another limitation of the PERSUADE corpus was that it was not clear that the models necessarily generalized to responses to unseen prompts. What we see from the distribution of responses to prompts, shown in Table \ref{tab:persuade-prompts}, is that each prompt is represented in the training sample. As mentioned above, this is a concern because, generally, the models used in scoring organization as a trait tend to be prompt-specific. Even in experiments in which a model is exposed to multiple prompts, the models seem to perform well on prompts from the training sample but poorly on unseen prompts.

\subsubsection{The AAE Dataset}

The AAE dataset was created using crowdsourcing, and the annotations were performed by human annotators. The scheme outlined by Stab and Gurevych specifies three argumentative components; a Major Claim (MC) that is typically (but not necessarily) expressed in both the introduction and conclusion outlining the author's position on a given topic, a set of Claims (Cl) that are arguments that support the Major Claim, and a set of Premises (Pr), which are facts that support the author's Claims. In addition to these argumentative components, the dataset also contains argumentative relations between the components. It is assumed that major claims form the root of any argument and that all other claims and premises either support or attack the major claim or other claims. This endows the argumentative components of an essay with the structure of a tree, which gives a very different understanding of argumentation.

The complete corpus of 402 essays, along with annotation guidelines, are freely available for download \footnote{Download available at \href{www.ukp.tu-darmstadt.de/data/argumentation-mining}{www.ukp.tu-darmstadt.de/data/argumentation-mining}}. This set is partitioned into a training set consisting of 322 essays and a test set consisting of 80 essays. The annotations above are applied to clauses, however, to simplify this situation, we apply these annotations at the word level. The relations and stances are considered binary labels that apply to pairs of components.

Given the above, and following the steps in \cite{stab_parsing_2017}, we can split the task of modeling into three stages:
\begin{enumerate}
\item {\bf Component identification:} This step identifies the boundaries of each component which in turn distinguishes argumentative text in an essay from non-argumentative text.
\item {\bf Component classification:} Once the boundaries of the argumentative components have been identified, this step classifies the component as being either a Major Claim, Claim, or Premise.  
\item {\bf Structure identification:} This step determines whether the argumentative components support or attack other argumentative components. 
\end{enumerate}

We start with the Argument Component Identification which is modeled using an BIO-tagset \cite{ramshaw_text_1995}. Each word in an BIO-tagset represents either the beginning of an annotation, given by B, the inside of an annotation, given by I, or outside the set of annotations, given by $O$. This means we treat this as a word-classification task in which every word is assigned a label. For every $B$ element, there is an associated component that requires classification as either a Major Claim (MC), Claim (Cl), or Premise (Pr). 

Every paragraph contains argumentative elements $(c_1, \ldots, c_n)$. The Relational identification data considers every pair $(c_i,c_j)$ for $i\neq j$ as a possible relation. This includes the possibility any argumentation supporting any other argumentative relation, even though Claims and Major Claims cannot support Premises. This defines a large collection of pairs of components that may or may not be linked. Since really only Premises can support and attack other components, this modeling only makes sense if one was to consider the argument relations as being independent of the component classification.

Lastly, we need to know the every stance. There are two types of stance that need to be considered; the stance of each Claim (Cl) and the stance of every Premise (Pr) that is linked to another argumentative component. This means that the stance data is explicitly dependent on the Relation identification and Argument Component Classification.

In summary this means for every essay we obtain 
\begin{itemize}
\item{A collection of IOB-labels for each word in the essay.}
\item{A component class for each element labeled B in the essay.}
\item{A possible link between each argumentative component in each paragraph in the essay.} 
\item{A stance for each Claim (Cl) and link in the essay.}
\end{itemize}
This gives us data associated with the training and test sets. A summary of this data is presented in Table \ref{tab:aae_summary}.

\begin{table}
\begin{tabular}{c|r r r | r r r | r r | r r} \toprule
\rowcolor{G4}& \multicolumn{3}{c|}{IOB} & \multicolumn{3}{c|}{Component} & \multicolumn{2}{c|}{Relation} & \multicolumn{2}{c}{Stance}\\
\rowcolor{G4}& \multicolumn{3}{c|}{Tagging} & \multicolumn{3}{c|}{Classification} & \multicolumn{2}{c|}{Identification} & \multicolumn{2}{c}{Recognition}\\
\rowcolor{G3} &&&&&&&\multicolumn{1}{c}{Not} &&&\\ 
\rowcolor{G3}          &  B & I  & 0 & MC & Cl & Pr & Linked & Linked & Support & Attack\\ \midrule
\rowcolor{G2}   Train  & 4,823 & 75,053 & 38,071 & 598 & 1,202 & 3,023  & 14,227 & 3,023 & 3,820 & 405 \\
\rowcolor{G1}       \%   & 4.1 & 63.6 & 32.3 & 12.4 & 24.9 & 62.7 & 82.5 & 17.5 & 90.4 & 9.6 \\
\rowcolor{G2}   Test   & 1,266 & 18,655 & 9,403 & 153 & 304 & 809 & 4,113 & 809 & 1,021 & 92\\
\rowcolor{G1}        \% & 4.3 & 63.6 & 32.1 & 12.1 & 24.0 & 63.9 & 83.5 & 16.5 & 91.7 & 8.3\\ \bottomrule
\end{tabular}
\caption{A summary of the training and test data used for the modeling of the AAE dataset.}
\label{tab:aae_summary}
\end{table}

\subsection{modeling details}

We start our modeling details with a discussion of the XLNet model. We state why it is different from many alternatives and why it is a more appropriate choice for modeling argumentation. We then discuss how we model each dataset, which includes how we chose to format the inputs into the model.

\subsubsection{The XLNet model}

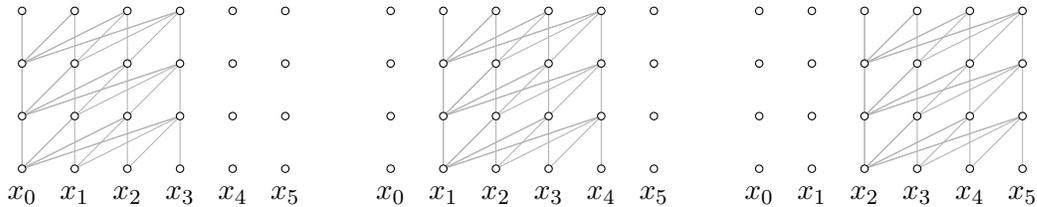
\begin{figure*}[!ht]
     \centering
     \begin{tikzpicture}[scale=0.7]
     \foreach \x in {0,8,16}
     {
         \foreach \y in {0,1,2}
         {
             \draw[black!30] (0+\x,\y) -- (0+\x,\y+1);
             \draw[black!30] (0+\x,\y) -- (1+\x,\y+1);
             \draw[black!30] (0+\x,\y) -- (2+\x,\y+1);
             \draw[black!30] (0+\x,\y) -- (3+\x,\y+1);
             \draw[black!30] (0+\x,\y) -- (0+\x,\y+1);
             \draw[black!30] (0+\x,\y) -- (1+\x,\y+1);
             \draw[black!30] (0+\x,\y) -- (2+\x,\y+1);
             \draw[black!30] (0+\x,\y) -- (3+\x,\y+1);
             \draw[black!30] (1+\x,\y) -- (1+\x,\y+1);
             \draw[black!30] (1+\x,\y) -- (2+\x,\y+1);
             \draw[black!30] (1+\x,\y) -- (3+\x,\y+1);
             \draw[black!30] (2+\x,\y) -- (2+\x,\y+1);
             \draw[black!30] (2+\x,\y) -- (3+\x,\y+1);
             \draw[black!30] (3+\x,\y) -- (3+\x,\y+1);
         }
     }
     \foreach \x in {0,1,2,3,4,5}
     {
         \foreach \y in {0,1,2,3}
         {
             \foreach \i in {0,7,14}
             {
                 \draw[draw=black,fill=white] (\x+\i,\y) circle (2pt);
             }
         }
         \foreach \i in {0,7,14}
         \node at (\x+\i,-.5) {$x_{\x}$};
     }
     \end{tikzpicture}
     \caption{The evaluation phase for the XLNet with a segment length of 4. This shows how the memory for a given segment is cached and used to extend the context for the next token.}
     \label{fig:recurrence}
 \end{figure*}

The transformer-based architectures defined by \cite{vaswani_attention_2017} have a fixed-length context. The novel contribution of the Transformer-XL, in \cite{dai_transformer-xl_2019}, was to introduce a recurrence mechanism within the architecture. The model is applied in segments where the hidden state of the previous segment is cached and reused as an extended context for the next segment. In this way, the hidden states are used as a memory state allowing for long-term dependencies that span beyond a single segment in a similar manner to a recurrent neural network. A diagrammatic representation of this recurrence is presented in Figure \ref{fig:recurrence}. This recurrence mechanism was also implemented in XLNet models \cite{yang_xlnet_2019}. In addition to this, XLNet and Transformer-XL use a relative positional embedding instead of an absolute positional embedding. As a result, Transformer-XL and XLNet do not have a maximum token limit. 

The main parameters defining any model are the number of layers, $N$, the number of heads, $h$, the dimension of the hidden layers, $d$, and the segment length, $L$. The recurrence relation determining how the hidden states are to be updated can be described as follows: suppose any input sequence of length $L$ is denoted $s_\tau = [x_{\tau,1}, \ldots, x_{\tau,L}]$ while the hidden state for $n$-th layer associated with $s_{\tau}$ is $h_{\tau}^n \in \mathbb{R}^{L\times d}$. The recurrence relation defining $h_{\tau+1}^n$ as a function of $h_{\tau}^{n-1}$ and $h_{\tau+1}^{n-1}$ is given as follows:
\begin{subequations}\label{eq:recurrence}
\begin{eqnarray}
\tilde{h}_{\tau+1}^{n-1} &=& [SG(h_{\tau}^{n-1}) \circ h_{\tau+1}^{n-1}] \\
q_{\tau+1}^{n}, k_{\tau+1}^{n}, v_{\tau+1}^{n} &=& h_{\tau+1}^{n-1} W_q, \tilde{h}_{\tau+1}^{n-1} W_k, \tilde{h}_{\tau+1}^{n-1} W_v\\
h_{\tau+1}^{n} &=& \mathrm{Transformer Layer}(q_{\tau+1}^{n}, k_{\tau+1}^{n}, v_{\tau+1}^{n}).
\end{eqnarray}
\end{subequations}
where $SG$ is the stop gradient and $[x\circ y]$ is the concatenation operation of two sequences. As in typical attention, the $W$ matrices are model parameters. It is important to note that this means that $h_{\tau+1}^n$ does not depend on $h_{\tau}^n$. This also means that the dependence on $h_{\tau}^{n-1}$ is limited to the keys and values, and not in the queries themselves. Note that, as mentioned above, $h_{\tau}^{n-1}$ is cached from the previous segment in this calculation.

Given how attention is defined and \eqref{eq:recurrence}, we find that there are possible dependencies between inputs of distance $L$ between two consecutive layers. This means that the maximal possible dependency length between two input tokens is $N \times L$ \cite{dai_transformer-xl_2019}, and hence, the maximum dependency grows linearly with the number of layers. While this is a theoretical limit, the implication is that deeper networks should handle long-term dependencies better. The paper on TransformerXL goes into some detail into the concept of a Relative Effective Context Length (RECL). The work suggests that this recurrence relation is far more effective at facilitating long-term dependencies that their recurrent unit counterparts, such as Long Short Term Memory (LSTM) units \cite{hochreiter_long_1997} and Gated Recurrent Unit (GRU) networks \cite{cho_learning_2014}.

The last advantage we wish to highlight in terms of XLNets exceptional suitability is the fact that XLNet uses permutation language modeling \cite{uria_neural_2016}. The XLNet models are essentially tuned in a similar manner to a masked language model, however, the number of outputs for the final linear layer is the number of annotation tags. The key idea is that models like BERT are trained by maximizing the loglihood function associated with the sequential prediction of masked tokens, whereas permutation modeling seeks to consider the interdependence between masked tokens by approximating the sum of the loglikihood function over all permutations using sampling. From the perspective of argumentation annotations, this approach considers the interdependence between annotation tags. 

There are two pretrained XLNet models available for fine-tuning; a base model and a large model \cite{yang_xlnet_2019}. The specifications for these models follow the base and large versions of BERT very closely \cite{devlin_bert_2019}. The specifications are listed in Table \ref{tab:pretrained_versions}. These two models were trained on the BookCopus \cite{zhu_aligning_2015}, Wikipedia, Giga5, ClueWeb, and Common Crawl. 

\begin{table}[ht]
    \centering
    \begin{tabular}{c|c c c c c} \toprule
\rowcolor{G4}        & Parameters & $N$ & $h$ & $d$ & $L$\\ \midrule
\rowcolor{G2}       Base  & $1.16\times10^8$  & 12 & 12 & 768 & 512 \\
\rowcolor{G1}       Large  & $3.6\times10^8$ & 24& 16 & 1024 & 512 \\ \bottomrule
    \end{tabular}
    \caption{The specifications for the two available pretrained versions of the XLNet model. The number of layers is $N$, the dimension of the each layer is $d$, the heads, given by $h$, refers to the number of attention heads, and $L$ refers to the segment length.}
    \label{tab:pretrained_versions}
\end{table}

When training our models, we used the well-defined train-test split for the PERSUADE corpus and AAE dataset and a random 10\% of the training set as a development set. This set was used to determine when to stop training the models, which was done after 20 epochs. Because we were optimizing for multiple metrics, the stopping condition was based on the sum of the metrics on the development set. All results shown are from the test set, and no results from the development set are reported.

XLNet does not have a formal maximum token limit due to its recurrent formulation. However, there are still hardware limitations. All models were trained on an NVIDIA RTX 8000 with 48GB of video memory using 16bit floats. We used variable-length inputs with a batch size of 1. Because of the size of the model, we truncated responses to 2048 tokens during training. We used Adafactor as an optimizer to save memory \cite{shazeer_adafactor_2018}. Learning rates varied by task. Lastly we note that one of the techniques that greatly improved our relevant accuracy statistics was to separate paragraphs using separator tokens during modeling. For this reason, we will see this in the form in the modeling details below for each dataset we modeled.

\subsubsection{ARROW Modeling}

The essays curated for the ARROW scheme were available in HTML format. We used the HTML paragraph definitions to split the essays into paragraphs, and then stripped away any remaining formatting. We used the SpaCy library to tokenize each paragraph into sentences. Since the annotations were applied at the sentence level, we also classified each sentence. The model input consisted of a mask token at the start of each sentence, and a separator between each paragraph. This means the input and targets for the model were as follows:
\begin{verbatim}
input = <mask><sentence 1 encoding>
        <mask><sentence 2 encoding>
        ...
        <mask>sentence p encoding><sep>
        <mask><sentence p+1 encoding>
        ...
        <sep><cls>
targets = <target 1><-> .... <->
          <target 2><-> .... <->
          ...
          <target p><-> .... <-><sep>
          <target p+1><-> .... <->
          ...
          <-><->
\end{verbatim}
where the \verb_<->_ is a token identified by the loss function as an element to be excluded in the loss calculation. This is the same loss calculation as masked language modeling so that the targets are calculated interdependently

The model is trained to assign a label to the mask preceding the sentence, and no other tokens. The above input and target means that there is a one-to-one mapping between sentence labels and mask tokens. The target for the mask token is assigned to the target tag for the sentence, while all other tokens are assigned a target token that is identified by the loss function as an element to be excluded in the loss calculation.

One of the key considerations in our modeling process was to ensure that the models would generalize to prompts that were not in the training sample. We knew that language models can be fine-tuned to assess language conventions and organization across several prompts, but that they are not as good at generalizing to new prompts when it comes to organization. This suggests models that assess organization tend to be specific to the prompts they were trained on. We took this into account when we designed our modeling process.

To assess whether our annotation scheme would generalize to prompts not used in the training sample, we trained five different models. Each model was trained on a different subset of the training sets for the nine prompts. We used one of the prompts as a development set and the double-scored validation set as a test set. We trained each model for 20 epochs and chose the model with the highest sum of kappa values across all the annotation tags on the development set. 

Once the five models were trained and inspected for quality, they were each applied to our large corpus of 250,000 of essays. This means that for each sentence, we had five predicted tags, one for each model. In most cases, there was a single most frequent predicted tag. However, when two or more tags were predicted with equal highest frequency, we assigned the tag in accordance with the hierarchy defined for the hand-scorers. A single universal model was trained on this large corpus and then the results were reported on the double-scored test set. In this way, we have used a large corpus essays with synthetic labels derived from models that have not been exposed to the final test set. 

\subsubsection{PERSUADE Modeling}

While the PERSUADE corpus is very similar in size and in nature to the corpus annotated using the ARROW scheme, a key difference is that it was annotated at the word level rather than the sentence level. In the context of the PERSUADE corpus, the set of words are given by sequences of non-space characters. In this way, we identify an essay as a collection of words rather than sentences, given by $(w_1, \ldots, w_n)$. Each word is then broken up into subwords according to XLNet's vocabulary of subword tokens. Target tags are aligned with the first word of each subword, and paragraphs are separated by separator tokens. 

In this formulation, consider the first $p$ words to belong to the first paragraph with word tokenizations \verb_<subword 1,0> ... <subword 1,n1>_ to \verb_<subword p,0> ... <subword p,np>_ then the appropriate form for the inputs and targets is given by
\begin{verbatim}
input = <subword 1,0> ... <subword 1,n1>
        <subword 2,0> ... <subword 2,n2> 
              ...
        <subword p,0> ... <subword p,np><sep>
        <subword p+1,0> ... <subword p+1,n(p+1)> ...
        ...
        <sep><cls>
targets = <target 1><-> ... <-> 
          <target 2><-> ... <-> 
                ...
          <target p><-> ...  <-><->
          <target p+1><-> ...  <->
          ...
          <-><->
\end{verbatim}
where the \verb_<->_ is a token identified by the loss function as an element to be excluded in the loss calculation. That is to say, we take a typical encoding of the essay, separating paragraph with separator tokens, and align targets with the first subword in the tokenization of each word while all other tokens are ignored in the loss calculation. 

\subsubsection{AAE Modeling}

In order to make our results comparable to those in \cite{stab_parsing_2017}, we follow similar modeling practices. This means that we model several key pieces of information in the AAE dataset with separate models. We have a total of four models for each pretrained version of XLNet. 

To model the BIO-tagset using XLNet, we know that every word is tokenized into possibly multiple subtokens. In a similar manner to the PERSUADE corpus, we treat this as a token-classification where the first subtoken of each word is assigned a label, in this case either $B$, $I$, or $O$, while all other tokens are assigned a null tag to indicate exclusion from the loss function. 

Once the Argument Component Identification model identifies a set of components, each component can be assigned a classification using the Argument Component Classification model. There are two ways to model this classification:
\begin{itemize}
\item{Assign annotation targets to the tokens with B-tags in the IOB-tagset either an MC, Pr, or Cl. This means that each token that is labeled as a B-tag is assigned the classification of the annotation target.}
\item{Attempt to classify each word in a block a tag and assign the block the tag that appears most frequently. This means that the first subtoken of each word that is labeled as an I or B is assigned either MC, CL, or Pr, and the component is then assigned the label that appears most frequently in the block.}
\end{itemize}
In our experiment, we found that the latter approach is more accurate. This is because averaging over the probabilities assigned over an entire argumentative component seems to provide more robust results. In this way, we assign each element designated an I or B with either an MC, Cl, or Pr, making three classes. All elements designated an O-tag are assigned tokens that indicate their exclusion in the loss calculation.

The Argument Relation Identification can be modeled by using XLNet as a sequence classifier. We found that the modeling relations was far more successful when the context for the relation was included in the model input. Relations occur are considered when argumentative components are single paragraph, hence, we augmented paragraph to include the tags \verb_<Source>_ and \verb_<Target>_ with separator tokens to indicate the location of the source and target of the linked components. This means that the input for our Argument Relation Identification model is of the following form: 
\begin{verbatim}
input  = ...  <sep><Source>:<arg1 encoding><sep> ...
         ...  <sep><Target>:<arg2 encoding><sep> ...
         ...
         <cls><sep>
\end{verbatim}
The target, in this case, is boolean variable indicating whether the two argumentative components are linked or not. 

Lastly, the Stance Identification model considers all relations between the claims and the major claims and the set of all links between argumentative components. The input for this model uses the same form of augmented paragraph used above for arguments linked within the same paragraph, and for any Claim (CL), this inlcuded the paragraph containing the claim and any paragraphs containing the Major Claim (MC). The target for this text classification task is the boolean variable that indicates whether the relation is supporting or attacking. In a similar manner, modeling using the full context of the relations was more accurate than modeling using only the argumentative text. 

\section{Results}

There are two aspects of the results; the performance of the models that identify argumentation components, and the implied relations between the tags across annotation schemes.

\subsection{Model performance}

In this section we show that the pretrained XLNet-models perform either above human baselines. 

\subsubsection{ARROW model}

We start with the five different models that were trained on the original hand-scored data curated for this study. These models are indexed by the prompts set aside as test sets. 

\begin{table}[!ht]
    \centering
    \begin{tabular}{c|c c c c c | c c c c c} \toprule
\rowcolor{G4}            & \multicolumn{5}{c| }{XLNet-Base} & \multicolumn{5}{c }{IRR} \\
\rowcolor{G3}           & 1 & 2 & 3 & 4 & 5 & 1 & 2 & 3 & 4 & 5\\ \midrule
\rowcolor{G2}        I1 & 0.805 & 0.696 & 0.770 & 0.761 & 0.740 & 0.805 & 0.690 & 0.785 & 0.746 & 0.711 \\
\rowcolor{G1}        I2 & 0.776 & 0.781 & 0.577 & 0.802 & 0.771 & 0.778	& 0.766 & 0.725 & 0.772 & 0.773\\
\rowcolor{G2}        E1 & 0.668 & 0.750 & 0.634 & 0.727 & 0.751 & 0.650 & 0.650 & 0.570 & 0.641 & 0.642 \\
\rowcolor{G1}        E2 & 0.641 & 0.661 & 0.627 & 0.680 & 0.728 & 0.624 & 0.588 & 0.593 & 0.644 & 0.602 \\
\rowcolor{G2}        O & 0.487 & 0.589 & 0.285 & 0.561 & 0.451 & 0.562 & 0.461 & 0.431 & 0.598 & 0.471\\
\rowcolor{G1}        C & 0.808 & 0.812 & 0.841 & 0.807 & 0.908 & 0.810 & 0.793 & 0.832 & 0.771 & 0.819\\
\rowcolor{G2}        T & 0.382 & 0.310 & 0.264 & 0.384 & 0.657 & 0.418 & 0.337 & 0.327 & 0.393 & 0.505\\ \midrule
\rowcolor{G3}       Avg & 0.697 & 0.715 & 0.622 & 0.723 & 0.725 & 0.705 & 0.658 & 0.656 & 0.695 & 0.670 \\ \bottomrule
    \end{tabular}
    \caption{The kappa statistics for the five models trained on the original dataset used in this study where each model is indexed by the test set used to evaluate the models.}
    \label{tab:seed_results}
\end{table}

With only a few exceptions, the seed models demonstrate remarkable performance. The inter-rater kappa values differ from the models' kappa values by no more than -0.1, a standard AES threshold \cite{williamson_framework_2012}, except for two values. On average, the models' kappa value is 0.02 higher than the inter-rater agreement across all prompts and annotation elements. The only two elements where the human raters outperformed the models were the Controlling Idea (I2) and Opposing Position (O). The model seemed to distinguish Elaboration (E1) and Evidence (E2) much more accurately than the human raters.

\begin{table}[!ht]
    \centering
    \begin{tabular}{p{4cm}|| l | l | l } \toprule
\rowcolor{G4}        & \multicolumn{1}{c|}{XLNet-Base} & \multicolumn{1}{c|}{XLNet-Large} & \multicolumn{1}{c}{H1-H2} \\
\rowcolor{G3}        Element              & $\kappa$ & $\kappa$ & $\kappa$  \\ \midrule
\rowcolor{G1}        Introduction         & 0.757 & 0.754 & 0.755\\
\rowcolor{G2}        Controlling Idea     & 0.770 & 0.768 & 0.735\\
\rowcolor{G1}        Evidence             & 0.683 & 0.683 & 0.627\\
\rowcolor{G2}        Elaboration          & 0.646 & 0.645 & 0.615\\ 
\rowcolor{G1}        Opposing Position    & 0.531 & 0.531 & 0.481\\
\rowcolor{G2}        Conclusion           & 0.832 & 0.832 & 0.828\\
\rowcolor{G1}        Transitions          & 0.428 & 0.423 & 0.429\\ 
\rowcolor{G2}        None                 & 0.405 & 0.437 & 0.415\\ 
\rowcolor{G3}        \midrule  Average              & 0.632 & 0.634 & 0.611\\ \bottomrule
    \end{tabular}
    \caption{The final model performance for the base and large XLNet models, presented in terms of both the $F_1$ score and cohen's weighted kappa statistic. This is compared with the agreement between two human raters.}
    \label{tab:final_model}
\end{table}

The results of Table \ref{tab:seed_results}, in comparison with the IRR values in Table \ref{tab:irr_handscoring}, indicate that the agreement between each model and the human resolved score is above the agreement between the raters used. This indicates that the models are at least as reliable as humans on an unseen prompt and that the resolution process applied to five models should, in theory, produce a sufficiently accurate implementation of the annotation scheme on an unseen prompt.  

The performance XLNet models trained on synthetic data is shown in Table \ref{tab:final_model}. The $F_1$ scores were added here to compare the agreement levels with the PERSUADE data. We see that these produce excellent results against their human benchmarks. In particular, the only elements the base model seems to have more trouble distinguishing are transitions and the untagged region. The large model produces excellent accuracy all around. 

\subsubsection{PERSUADE model}

Given the success of XLNet on our own annotation scheme, it is natural to ask how well XLNet applies to the PERSUADE corpus. In order to make such a comparison, we need to alter the modeling very slightly to apply to a word-level annotation. Note that the words in the PERSUADE corpus are simply defined as strings separated by a space. Given a word, we use the tokenization of XLNet to divide each word into a collection of subwords. In the training phase, we apply labels to the first subword and ignore the remaining subwords in the calculation of the loss function. We isolated a random 10\% sample of the training set as a development set where we chose the best performing model in accordance with the $F_1$ score using macro-averaging. Apart from these changes, we subjected the models to the training regime specified in the methods section. The $F_1$ values for both model trained from the base and large pretrained XLNet models can be found in Table \ref{tab:my_label}.

Since which elements of the public and private leaderboard have not been disclosed, we simply report scores at the individual token level, as defined in the competition by the split function. This is a stricter condition than allowing for 50\% overlap. 
\begin{table}[!ht]
    \centering
    \begin{tabular}{p{2cm}|| l l  } \toprule
\rowcolor{G4}    & \multicolumn{1}{c}{XLNet-Base} & \multicolumn{1}{c}{XLNet-Large}\\
\rowcolor{G3}        Element & F1  & F1 \\ \midrule
\rowcolor{G2}        L & 0.800 & 0.811  \\
\rowcolor{G1}        P & 0.696 & 0.713  \\
\rowcolor{G2}        C1 & 0.529 & 0.555  \\
\rowcolor{G1}        C2 & 0.504 & 0.553 \\ 
\rowcolor{G2}        R & 0.418 & 0.467  \\
\rowcolor{G1}        E & 0.717 & 0.731  \\
\rowcolor{G2}        C3 & 0.837 & 0.837 \\ \midrule
\rowcolor{G3}        Macro Avg & 0.642 & 0.667\\ \bottomrule
    \end{tabular}
    \caption{The inter-rater reliability statistics for each annotation label for the base and large XLNet models.}
    \label{tab:my_label}
\end{table}
It should be noted that the only other published result seems to be an F1 score of 0.63 reported in \cite{ding_score_2023}.

The performance of the large XLNet model is very strong for a single model (not an ensemble). While few benchmarks have been published on the entire test set, there are many results available on the Kaggle discussion boards on the private and public splits of those sets. In particular, there are many results in which the essays are processed by splicing them into segments of lengths that can be processed by language models \footnote{See solutions posted on https://www.kaggle.com/ competitions/feedback-prize-2021/leaderboard}. Given the unknown partition of the test set into public and private, we do not know if these are comparable.

What we can say about the approaches is that most entries account for the length limitation is taken into account by partitioning texts into multiple segments. Secondly, that many of the top results include ensembling models like DeBERTa over multiple folds, removing small segments, employing variable cutoffs and using hyperparameter tuning, all of which are novel ideas that would surely improve the results. Given our task is to prove the effectiveness of XLNet on its own, we do not pursue these optimizations.

\subsubsection{AAE model performance}

The AAE modeling is different from the above in that the process, as described in \cite{stab_annotating_2014}, is broken up into four distinct modeling tasks, meaning that there are four different models to consider; an IOB-tagger, a classification model, a stance model, and a link model. In each case, the model performance is measured by the F1-scores for each of the tags. Baseline scores are those reported in \cite{stab_parsing_2017}. 

\begin{table}[]
    \begin{small}
    \centering
    \begin{tabular}{c|c c c c | c c c c | c | c  } \toprule
\rowcolor{G4}         & \multicolumn{4}{c|}{Components} & \multicolumn{4}{c|}{Components} & \multicolumn{1}{c}{Rels} & \multicolumn{1}{|c}{Stance} \\
\rowcolor{G3}          &  F1 & F1 B & F1 I & F1 O & F1 & F1 MC & F1 Cl & F1 Pr & F1 & F1   \\ \midrule
\rowcolor{G2}        Human & 0.886 & 0.821 & 0.941 & 0.892 & 0.868 &  0.926 & 0.754 & 0.924 & 0.854 & 0.844 \\
\rowcolor{G1}     Results of \cite{stab_parsing_2017} &  0.867 & 0.809 & 0.934 & 0.857 & 0.826  & 0.891 & 0.682 & 0.903 & 0.759 & 0.702 \\
\rowcolor{G2}         XLNet Base & 0.936 & 0.956 & 0.951 & 0.947 & 0.847 & 0.924 & 0.814 & 0.950 & 0.857 & 0.827\\ 
\rowcolor{G1}        XLNet Large &  0.933 & 0.954 & 0.951 & 0.946 & 0.870 & 0.932 & 0.846 & 0.940 & 0.896 & 0.885\\ \bottomrule
    \end{tabular}
    \end{small}
    \caption{The reuslts from \cite{stab_parsing_2017} include a Conditional Random Field (CRF) to determine BIO tags, Integer Linear Programming (ILP) models to identify components and relations, and an SVM to classify stance. The XLNet models are either token classifiers for the BIO tagging and Component classifiers or sequence classifiers in the case of the Relation and Stance classification. \label{tab:xlnet-stab}}
    \label{tab:xlnet-stab}
\end{table}

We listed the a summary of F1 scores for each of the models in Table \ref{tab:xlnet-stab}. In this table, we included the results of \cite{stab_parsing_2017} and the human benchmarks. The results showed that the XLNet models outperformed previous models on all of the tasks, and even achieved performance that is comparable to, and in many cases above, human benchmarks.

It should be noted that the modeling can be done in many different ways. For example, instead of IOB-tagging, one could try to predict the labels themselves as one consistent block, or predict endpoints of each argument. As with the case of the PERSUADE corpus, there are many optimizations that could be applied to this modeling process.

\subsection{Relations between the Annotation Schemes}

The annotation schemes described in the previous sections reflect different perspectives on argumentation. Each scheme defines the role of different parts of an essay in establishing a coherent argument. Since each scheme segments arguments into components, there should be relationships between the tags used by the different schemes. The models presented in this study provide approximations of how each annotation scheme applies to the different datasets used. These approximations allow us to explore the possible correspondences between the annotation schemes.

\begin{figure}[!ht]
\centering
\fbox{\includegraphics[page=1,width=5.7in, trim=0.2in 2in 0.2in 1in]{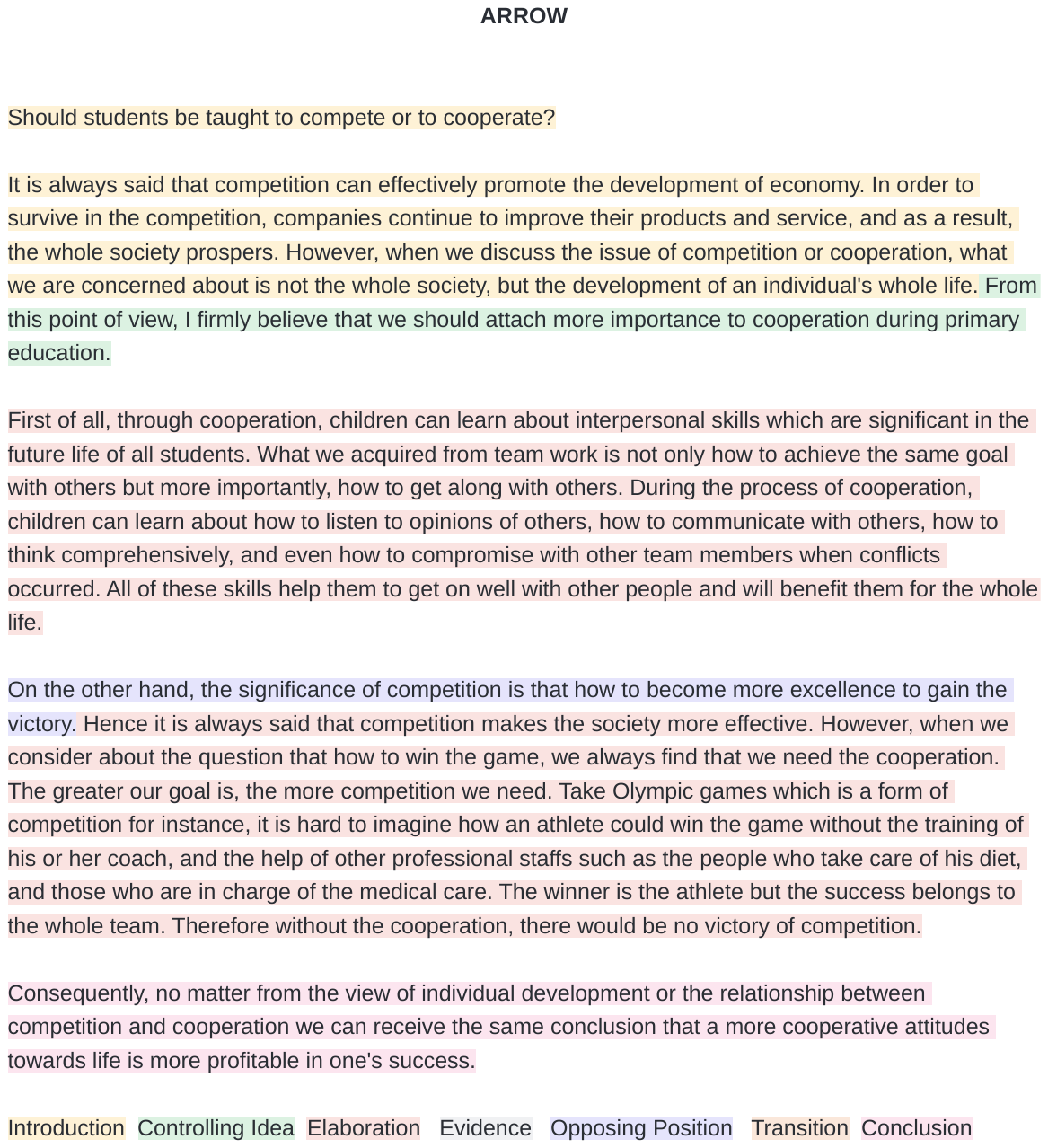}}
\caption{An essay from the AAE dataset that has been annotated with respect to the ARROW scheme. In this particular essay there were no sentences that were considered Transitions (T) or Evidence (E1). \label{fig:ARROW-essay1}}
\end{figure}

\begin{figure}[!ht]
\fbox{\includegraphics[page=1,width=5.7in, trim=0.2in 2in 0.2in 1in]{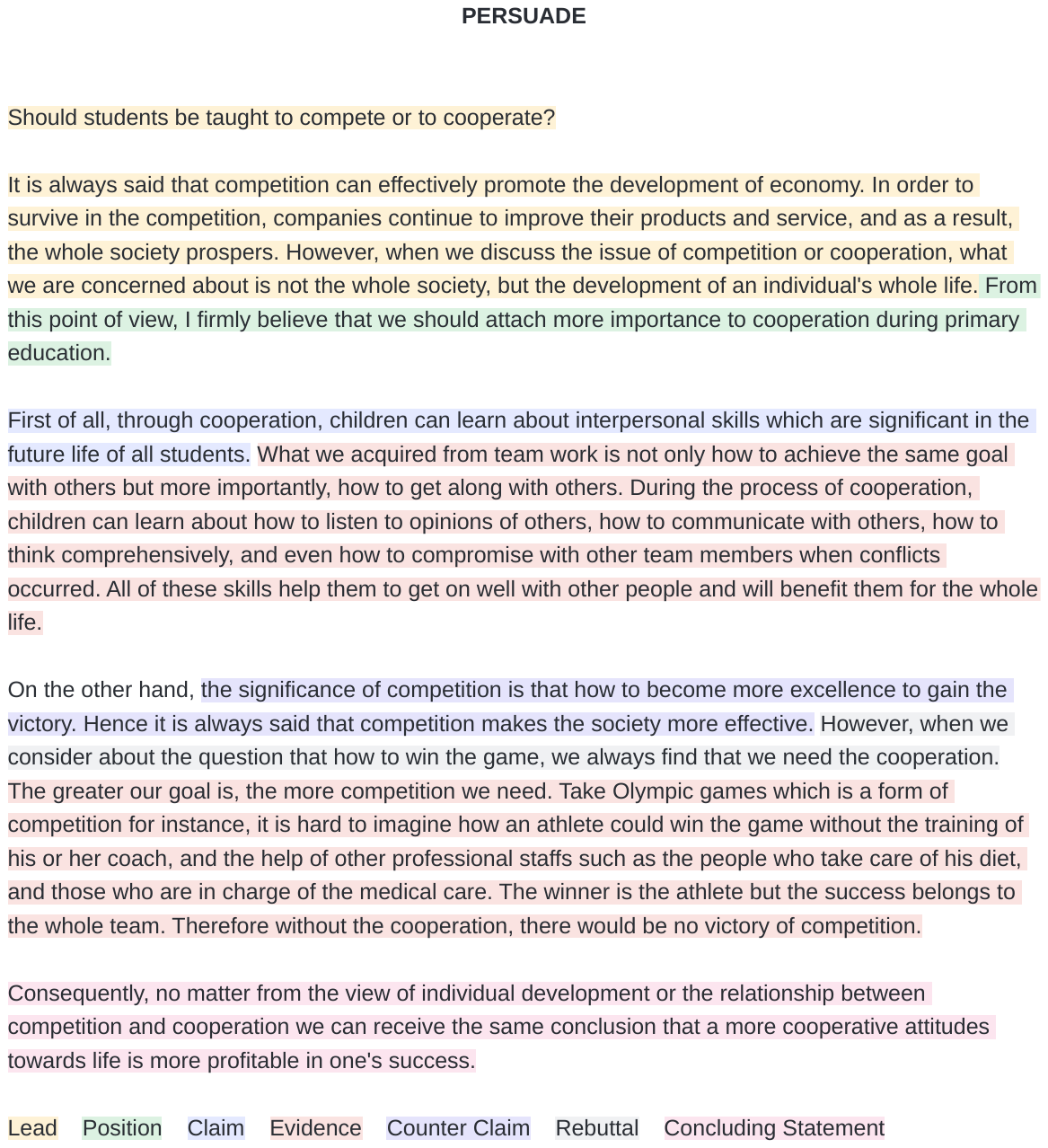}}
\caption{The same essay used previously from the AAE dataset annotated using the PERSUADE scheme. \label{fig:PERSUADE-essay1}}
\end{figure}

\begin{figure}[!ht]
\fbox{\includegraphics[page=1,width=5.7in, trim=0.2in 2in 0.2in 1in]{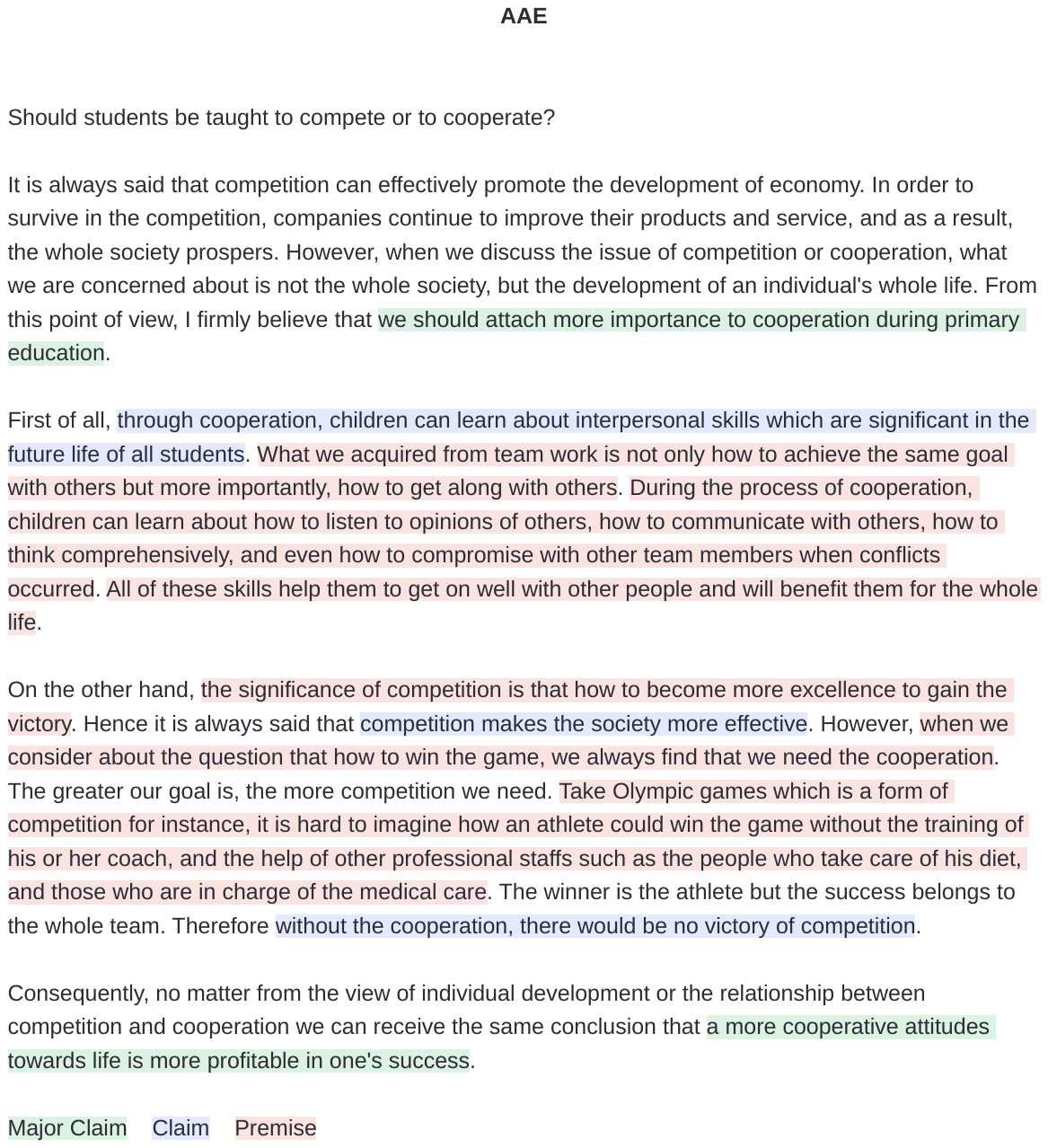}}
\caption{The same essay from the AAE dataset used in previous figures where the argumentative components of the essay were highlighed using the human assigned annotations. \label{fig1:AAE-essay1}}
\end{figure}

Our first demonstration is a visual one. We take a single essay and use the models to synthesize annotations from each of the schemes. We use the human-defined annotation tags in one version of the essay from the AAE dataset, and synthetic tags to approximate the application of the PERSUADE and ARROW schemes. Color-coded versions of the essay are presented in Figure \ref{fig:ARROW-essay1}, Figure \ref{fig:PERSUADE-essay1}, and Figure \ref{fig1:AAE-essay1}.


The other way in which correspondences may be inferred is to consider what percentage of the human annotations receive the various annotation tags from the other schemes. While we do not have the ability to get humans to annotate each of the datasets, we can use the models ability to approximate the application of the annotation scheme. To compare the schemes we collapse each annotation scheme to the word level. So a tag applied to a token in the PERSUADE dataset, where every token was separated by a space, would apply to every word separated by punctuation as well.  In terms of the ARROW scheme, the application of a tag to a single sentence would be applied to every word in that sentence instead. This gives us three datasets, each annotated at the word level where each word has a human annotation tag defined from the dataset itself, and two synthetic tags defined by the Large XLNet models. The percentage of synthetic annotations for each word given a particular human assigned annotation for each of the datasets above is presented in Table \ref{tab:arrow_correspondence}, Table \ref{tab:persuade_correspondence}, and Table \ref{tab:aae_correspondence}.

\begin{table}[!ht]
    \centering
    \begin{tabular}{c|r | r r r | r r r r r r r r} \toprule
\rowcolor{G4}    && \multicolumn{3}{c|}{AAE}& \multicolumn{7}{c}{PERSUADE}\\
\rowcolor{G3}         & \% &  MC & Cl & Pr            & L & P & C1 & C2 & R & E & C3\\ \midrule 
\rowcolor{G2}        I1 & 10.8  &  2.4& 14.5 & 35.3   & 40.0 & 7.3 & 20.2 & 2.2 & 1.3 & 32.3 & 0.0\\
\rowcolor{G1}        I2 & 5.6 &  26.9 & 19.1 & 18.1   & 6.9 & 48.7 & 17.2 & 0.6 & 0.6  & 12.2 & 6.7 \\
\rowcolor{G2}        E1 & 19.3 &  1.2 & 7.9 & 60.3    & 1.3 & 0.4 & 4.8 & 1.7 & 2.0 &  87.0 & 1.1\\
\rowcolor{G1}        E2 & 46.9 &  1.6 & 14.9& 62.7    & 0.9 & 1.0 & 11.9 & 2.2 & 4.3 & 73.3 & 3.3\\
\rowcolor{G2}        O & 5.8 &  1.2 & 15.6 & 53.8     & 0.6 & 0.3 & 4.6 & 27.2 & 10.3 & 53.3 & 2.4 \\
\rowcolor{G1}        C & 9.5 &  15.4 & 26.6 & 19.3    & 0.0 & 1.1 & 0.7 & 0.4 & 0.7 & 5.0 & 83.9 \\
\rowcolor{G2}        T & 0.9&  1.6 & 43.8  & 13.1     & 0.0 & 0.3 & 36.4 & 2.1 & 0.7 & 20.0 & 0.5\\ \midrule
\rowcolor{G3}         \%   & & 4.3 & 15.0 & 51.1        & 5.5 & 4.0 & 10.3 & 3.3 & 3.2 & 58.4 & 10.3\\ \bottomrule 
    \end{tabular}
    \caption{We present the percentage of annotation tags assigned by humans with respect to the ARROW scheme that were assigned each AAE tag and PERSUADE tag by the Large XLNet model. The rows are labeled by the human annotations, while the columns are labelled by the synthetic annotations applied by the models. }
    \label{tab:arrow_correspondence}
\end{table}

\begin{table}[!ht]
    \centering
    \begin{tabular}{c| r |r r r |  r r r r r r r r} \toprule
\rowcolor{G4}         && \multicolumn{3}{c|}{AAE}& \multicolumn{7}{c}{ARROW}\\
\rowcolor{G3}         & \% &  MC &  Cl&      Pr  & I1 &  I2 & E1 &  E2  &  O &  C &   T\\ \midrule
\rowcolor{G2}      L  & 6.5 & 2.6  &6.9  &24.3   & 82.6& 	8.3& 	2.3& 	6.5&	0.1&	0.1&	0.0 \\
\rowcolor{G1}      P  & 4.1 &34.1 &18.1 &8.5   & 22.5 &	59.0&	1.7&	10.8&	0.0&	5.8&	0.1 \\
\rowcolor{G2}      C1 & 12.6 &2.3  &35.6 & 42.8  &14.4&	10.5&	8.0&	62.7&	0.7&	1.0&	2.6 \\
\rowcolor{G1}      C2 & 2.0 &0.8  &25.2 &42.8   & 6.8&	2.5&	3.9&	44.3&	39.5&	2.8&	0.1 \\
\rowcolor{G2}      R  & 1.8  &2.9  &11.7  &65.4   &4.1&	1.5&	6.5&	72.2&	10.0&	5.6 & 0.1 \\
\rowcolor{G1}      E  & 48.8 &0.9  &9.7 &73.4   & 3.4& 1.1&	18.4&	74.1&	1.3&	1.3&	0.4 \\ 
\rowcolor{G2}      C3 & 11.5 &15.6  &29.4  &22.8 & 0.0&	1.0&	1.8&	14.3&	0.6&	82.3&	0.0 \\ \midrule 
\rowcolor{G3}      \% & & 4.4 & 15.9 & 51.0 & 11.7 & 5.8 & 12.2 & 55.5 & 1.9 & 11.7 & 1.2 \\ \bottomrule
    \end{tabular}
    \caption{We present the tags in the PERSUADE corpus where each row is labelled by the human annotation tags and the columns represent the ARROW and AAE synthetic labels using the XLNet-Large models. The entry in each row and column represents the percentage of the human tags of the row were assigned the synthetic tag of that column.}
    \label{tab:persuade_correspondence}
\end{table}

\begin{table}[!ht]
    \centering
    \begin{small}
    \begin{tabular}{c|r|r r r r r r r |  r r r r r r r} \toprule
\rowcolor{G4}    && \multicolumn{7}{c|}{PERSUADE}& \multicolumn{7}{c}{ARROW}\\
\rowcolor{G3}         &  &  L &  P&      C1  &  C2& E  &   R  & C3 &  I1 &  I2 & E1 &  E2  &  O &  C &   T\\ \midrule
\rowcolor{G2}      MC & 7.2  & 2.5 & 41.1 & 2.4 & 0.0 & 0.0 & 0.0 & 52.8 & 9.1 & 35.7 & 0.0 & 0.0 & 0.2 & 55.0 & 0 \\
\rowcolor{G1}      Cl & 15.0 & 1.2 & 2.3 & 37.9 & 6.8 & 30.0 & 1.9 & 18.7 & 4.5& 2.5 & 0.0 & 65.5 & 6.3 & 19.2 & 1.8\\
\rowcolor{G2}      Pr & 44.7 & 0.0 & 0.0 & 8.1 & 2.1 & 84.7 & 3.6 & 1.3 & 0.3 & 0.0 & 3.0 & 90.8 & 0.4 & 1.4 & 0.3\\ \bottomrule
\rowcolor{G3}      \% & & 11.1 & 6.0 & 11.0 & 48.6 & 2.9 & 2.5 & 11.8 & 15.1 & 5.9 & 1.6 & 60.4 & 3.7 & 12.6 & 0.6 \\ \bottomrule
    \end{tabular}
    \end{small}
    \caption{We present the tags in the AAE dataset where each row is labelled by the human annotation tags, whereas the columns represent synthetic labels from the PERSUADE and ARROW annotation tags that were applied using the XLNet-Large models. Each entry represents the percentage of the human annotations in that each row that were assigned each synthetic tag by the XLNet large models in each column.}
    \label{tab:aae_correspondence}
\end{table}

The percentage of words in each corpus that were considered Evidence (E) was highest for the ARROW corpus, followed by the PERSUADE corpus. We believe this is naturally because the prompts used in the ARROW corpus were all source-dependent, while the PERSUADE corpus consisted of essays responding to a mix of source-independent and source-dependent prompts. All responses in the AAE dataset were independent of a source text.

The dataset with the highest percentage of Major Claim (MC) was the AAE dataset. We believe this may be because the essays chosen for the AAE dataset needed to be of a certain quality in terms of spelling and length to be chosen for the study. The positive correlation between spelling and overall quality suggests that the Major Claim in an essay could have been clearer in the higher quality essays than they are in the poor-performing essays. This also means that the models trained to identify the AAE tags may not be as robust to poor spelling as those trained on the other datasets. Poor spelling and small training set size may have contributed to tagging sections with a lower model confidence, and hence, a more even distribution of AAE tags in Tables \ref{tab:arrow_correspondence} and \ref{tab:persuade_correspondence}.

There are some interesting correspondences between tags. For example, we see words assigned Major Claim (MC) tags are almost completely contained in those tagged with either the Position (P) and Concluding Statement (C3) in the case of the PERSUADE annotation scheme, or the Controlling Idea (I2) and Concluding Statement (C3) in the case of the ARROW scheme. We believe this correspondence is not as clear in Tables \ref{tab:arrow_correspondence} and \ref{tab:persuade_correspondence}, again, because of the poor spelling and small training set size.

Similarly, the tokens assigned the Premise tag (Pr) by humans are almost completely assigned Evidence (E) and Elaboration (E2) by our models. There are also very strong correspondences between Conclusion (C) and Concluding Statements (C3). It should also be noted that human-assigned Evidence (E1) and Elaboration (E2) have both predominantly been given the synthetic Evidence (E) tag by the models, suggesting that Evidence (E1) and Elaboration (E2) are essentially a partition of the Evidence (E).

\section{Discussion}

The persistent issue of managing lengthy essays and the associated long-term dependencies poses a challenge when employing language models for essay scoring, and this complexity extends to modeling annotation schemes. We contend that architectures extending the innovations of Transformer-XL and XLNet present a compelling approach to surmounting these length constraints. Although this consideration was part of the early discussions in Automated Essay Scoring (AES) initiatives \cite{rodriguez_language_2019}, the prevailing trend in current research on language modeling for AES involves the use of models with inherent length limitations \cite{uto_neural_2020, ormerod_automated_2021, yang_enhancing_2020}.

This study highlights notable distinctions between the ARROW system and the scheme employed in the PERSUADE corpus. Specifically, an examination of the predominant tags in each scheme, namely Evidence (E) and Elaboration (E2), reveals that ARROW adopts a more stringent definition of evidence, encompassing external sources like data or excerpts from source texts. This distinction is crucial in the context of assessment, as the capacity to cite evidence for argumentative support aligns with established standards across various grade levels. However, a limitation of ARROW lies in the fact that its sentence-level annotation schemes lack the granularity found in word-level schemes such as the PERSUADE corpus. The utilization of sentence-level annotations may introduce ambiguities, particularly when essays lack well-defined sentence boundaries. It is not uncommon, especially in lower grades, to encounter essays that essentially form a single extended sentence. Additionally, the practice of elaborating on a controlling idea and presenting evidence within a single sentence is prevalent. A potential enhancement could involve defining a more suitable approach in terms of elementary discourse units (EDU) or clauses, as proposed by Li et al. \cite{li_composing_2020}.

An additional constraint of this study stems from the reliance on models trained with the original dataset for generating synthetic data. Specifically, for the AAE dataset, we posit that incorporating adversarial training—exposing models to randomized spelling errors—could enhance their performance. The analysis of the ARROW dataset in this research offers compelling evidence supporting the idea that ARROW models, and by extension, PERSUADE models, exhibit transferability to prompts not encountered during training. This valuable insight would have been inaccessible if solely examining the PERSUADE set in isolation.

As highlighted by various methods outlined on the Kaggle website, there exist numerous strategies for improving the precision of annotating essay prompts. These approaches encompass hyperparameter tuning, employing variable thresholds for individual tags, and employing $k$-fold ensembling to optimize the utilization of the training set. In practical terms, our models were crafted to facilitate straightforward implementation in AWE systems, ensuring a seamless integration while upholding high accuracy.

\bibliographystyle{plain}
\bibliography{lib}{}

\end{document}